# CauTraj: A Causal-Knowledge-Guided Framework for Lane-Changing Trajectory Planning of Autonomous Vehicles

Cailin Lei, Haiyang Wu, Yuxiong Ji*, Xiaoyu Cai, and Yuchuan Du, Member, IEEE

*Abstract*—Enhancing the performance of trajectory planners for lane-changing vehicles is one of the key challenges in autonomous driving within human–machine mixed traffic. Most existing studies have not incorporated human drivers' prior knowledge when designing trajectory planning models. To address this issue, this study proposes a novel trajectory planning framework that integrates causal prior knowledge into the control process. Both longitudinal and lateral microscopic behaviors of vehicles are modeled to quantify interaction risk, and a staged causal graph is constructed to capture causal dependencies in lane-changing scenarios. Causal effects between the lane-changing vehicle and surrounding vehicles are then estimated using causal inference, including average causal effects (ATE) and conditional average treatment effects (CATE). These causal priors are embedded into a model predictive control (MPC) framework to enhance trajectory planning. The proposed approach is validated on naturalistic vehicle trajectory datasets. Experimental results show that: (1) causal inference provides interpretable and stable quantification of vehicle interactions; (2) individual causal effects reveal driver heterogeneity; and (3) compared with the baseline MPC, the proposed method achieves a closer alignment with human driving behaviors, reducing maximum trajectory deviation from 1.2 m to 0.2 m, lateral velocity fluctuation by 60%, and yaw angle variability by 50%. These findings provide methodological support for human-like trajectory planning and practical value for improving safety, stability, and realism in autonomous vehicle testing and traffic simulation platforms.

## I. INTRODUCTION

Lane-changing trajectory planning has gained significant attention in autonomous driving research. The performance of planned trajectories directly affects traffic flow stability and driving comfort [1]. In mixed traffic with both human-driven and autonomous vehicles, the interaction between lane-changing vehicles (LCVs) and surrounding vehicles (SVs) is dynamic. This brings greater challenges to autonomous driving systems [2, 3]. For example, autonomous vehicles need to consider the influence of SV behaviors to plan human-like trajectories. Therefore, in lane-changing scenarios, more attention should be given to understanding the causal relationships of microscopic interactions between vehicles and integrating the causal effects of human driving behavior into trajectory planning methods.

The core of vehicle trajectory planning is to move safely and efficiently from the current position to the target position. Existing research mainly follows two directions: deep learning (DL) and optimization [2, 4].

DL-based methods rely on large-scale driving data to learn policies for trajectory planning, aiming to imitate or surpass human driving. For instance, Zhu, et al. [1] applied reinforcement learning (RL) to naturalistic data to construct a human-like car-following model, while WANG, ET AL. [5] combined deep RL with rule-based constraints to improve lane-changing safety. Cheng, et al. [6] introduced a Monte Carlo tree search–based RL algorithm for safety in mixed traffic, and Chai, et al. [7] proposed a real-time trajectory planning and tracking framework for AGVs in complex parking. Zhang, et al. [8] formulated a partially observable Markov decision process (POMDP) with deep RL, outperforming humans in safety and efficiency, whereas Xu, et al. [9] designed a centralized-decision distributed-planning scheme to enhance efficiency in unstructured conflict zones. Despite strong potential, DL approaches suffer from limited interpretability [10-14], restricting their use in safety-critical tasks.

Optimization-based methods, in contrast, explicitly model objectives (e.g., safety, comfort, efficiency) and constraints (e.g., dynamics, collision avoidance, traffic rules), aiming for optimal or near-optimal trajectories. Multi-objective optimization, curve fitting, and model predictive control (MPC) are the main approaches. Zhao, et al. [15] proposed GDTP-RRT for long-distance planning with high accuracy, while Wang, et al. [16] and Liu, et al. [17] used polynomial interpolation to generate smooth, safe trajectories. Wu, et al. [18] developed a co-evolutionary method to reduce lane-change conflicts, and Möller, et al. [3] balanced safety, comfort, and tracking accuracy with a multi-objective algorithm.

Specially, model predictive control (MPC) has become a core technique in trajectory planning due to its capability of handling constraints and performing rolling optimization. It has attracted significant research attention. For example, Li, et al. [19] combined quintic polynomials with MPC to achieve optimal lane-changing trajectories by minimizing regional

This work was supported by the General Natural Science Fund of Chongqing Science and Technology Bureau (CSTB2025NSCO-GPX0902), Youth Project of Chongqing Municipal Education Commission (KJQN202500766), and The Leading Project of Chongqing Jiaotong University in Natural Sciences (XJ2023000801). (*Corresponding author: Yuxiong Ji*)

Cailin Lei and Xiaoyu Cai are with the School of Smart City, Chongqing Jiaotong University, Chongqing 400074, China (e-mail: cllei@cqjtu.edu.cn; caixiaoyu@cqjtu.deu.cn); Haiyang Wu is with the College of Traffic and Transportation, Chongqing Jiaotong University, Chongqing 400074, China (e-mail: 2010530510@qq.com); Yuxiong Ji and Yuchuan Du are with the Key Laboratory of Road and Traffic Engineering of the Ministry of Education, Tongji University, Shanghai 201804, China (e-mail: yxji@tongji.edu.cn; ycdu@tongji.edu.cn)

Corresponding author: Yuxiong Ji, yxji@tongji.edu.cn

costs. Qie, et al. [20] improved the MPC method by integrating Kalman filter-based fusion to predict obstacle trajectories and their uncertainties, thereby enhancing the robustness and stability of the planned trajectories. Ji, et al. [21] integrated ellipsoidal potential fields with Gaussian velocity fields (GVF) and developed an MPC-based trajectory planning method, which was validated in both lane-changing and car-following scenarios. Li, et al. [22] approximated non-convex obstacle avoidance constraints into convex forms and embedded them into the MPC framework. This approach enabled optimal trajectory planning under simultaneous longitudinal and lateral obstacle avoidance constraints, while ensuring both safety and comfort.

Existing studies on vehicle trajectory planning suggest that optimization-based methods offer better interpretability. They also show that incorporating the behavior of SVs is an effective way to improve the rationality of planned trajectories. Introducing prior knowledge into trajectory planning or decision-making can further enhance the performance of trajectory planners [22-24]. However, causal prior knowledge regarding the interactions between the ego vehicle and SVs has not yet been considered in trajectory planning.

To fill above gap, this study proposes a lane-changing trajectory planning framework that incorporates causal prior knowledge. Firstly, the lane-changing process is divided into stages, and risk-related causal factors are identified for each stage. Secondly, a causal graph is constructed to represent the relationships between these factors, and causal inference techniques are applied to estimate causal effects. Thirdly, MPC-based trajectory planning method is developed that integrates causal prior knowledge. Finally, the effectiveness of the proposed method is validated using vehicle trajectory data under naturalistic driving conditions.

The main contributions of this study are as follows:
(1) Causal effects between SVs and LCV behaviors are quantified for lane-changing scenarios.
(2) Causal prior knowledge is incorporated into the MPC framework to improve the performance of MPC-based trajectory planning.
(3) The proposed method is validated using naturalistic driving trajectory data.

The remainder of this paper is organized as follows. Section 2 presents the proposed methodology. Section 3 demonstrates the case study results. Section 4 provides discussion. Section 5 concludes the paper and outlines future research directions.

## II. METHODOLOGY

Fig. 1 illustrates the proposed lane-changing trajectory planning framework that incorporates causal prior knowledge, which consists of three main stages. Stage 1 describes the calculation of interaction risk between vehicles and the selection of causal factors. Stage 2 focuses on the computation of causal effects, including the construction of the causal graph and the estimation of causal effects. Stage 3 presents the lane-changing trajectory planning process based on the integration of prior causal effects and MPC.

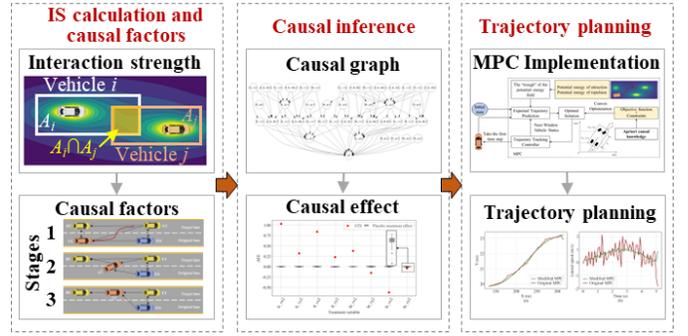

**Fig. 1** The framework of proposed in this study.

### A. Interaction Risk and Causal Factors
#### a). Interaction Risk

Ma, et al. [24] proposed that the interaction between vehicles can be represented by the overlapping part of potential energy field of those vehicles. Therefore, the proportion of area and potential energy in overlapping part was employed to represent the interaction strength (IS) between neighboring vehicles [25]. The higher IS value indicates a higher risk.

Fig. 2 show the overlap of influence regions for vehicle $i$ and $j$. The IIS for vehicle $i$ and $j$ is calculated by:

$$IS_{ij} = ROL_{i,j} \times (RE_i + RE_j)/2 \tag{1}$$

where the maximum and minimum of IIS are 0 and 1, respectively. The $ROL_{ij}$ is the area proportion of overlapping part of the influence regions for vehicle $i$ and $j$. $RE_i$ and $RE_j$ are potential energy proportion of overlapping part of the influence regions for vehicle $i$ and $j$, respectively. $ROL_{ij}$, $RE_i$ and $RE_j$ are calculated by:

$$ROL_{ij} = \frac{(x_{j1} - x_{i1}) \times (y_{j2} - y_{j1})}{\min\left[(x_{i2} - x_{i1}) \times (y_{i2} - y_{i1}), (x_{j2} - x_{j1}) \times (y_{j2} - y_{j1})\right]} \tag{2}$$

$$RE_i = \int_{y_{i1}}^{y_{j2}} \int_{x_{j1}}^{x_{i2}} E_i(x,y) dx dy \Big/ \int_{y_{i1}}^{y_{i2}} \int_{x_{i1}}^{x_{i2}} E_i(x,y) dx dy \tag{3}$$

$$RE_j = \int_{y_{i1}}^{y_{j2}} \int_{x_{j1}}^{x_{j2}} E_j(x,y) dx dy \Big/ \int_{y_{j1}}^{y_{j2}} \int_{x_{j1}}^{x_{j2}} E_j(x,y) dx dy \tag{4}$$

A more detailed calculation process for IS can be found in reference [25].

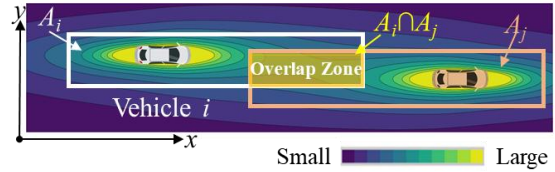

**Fig. 2** Schematic diagram of the overlap zone for two vehicles.

#### b). Causal Factors

In the inference of driving risk causes for LCVs, the key steps include the causal factors selection, causal graph construction, and causal effects estimation and validation [23, 26, 27]. This section introduces the selection of causal factors, whereas the following sections detail the construction of causal graphs and the estimation of causal effects.

Fig. 4(a) to 4(c) illustrate the influence relationships between vehicles during the three stages of a lane-change. LC denotes the LCV, FO denotes the front vehicle on the original lane, FT denotes the front vehicle on the target lane, and BT denotes the behind vehicle on the target lane. Following the stage division of lane-change maneuvers [28-31], this study divides the lane-change process into three stages. Specifically, Stage 1 refers to the 2 s period before the LCV generates lateral speed. Stage 2 covers the period from the initiation of the lane-change to the complete insertion into the target lane. Stage 3 refers to the 1 s period after the vehicle has fully merged into the target lane.

In the pre-lane-change stage, the driver observes the behavior and spatial positions of FO, FT, and BT. During the lane-change process, the driver continues to be influenced by the behavior and distances of these vehicles. Therefore, after the maneuver, the interaction risk is mainly shaped by the front and behind vehicles on the target lane as well as the front vehicle on the original lane.

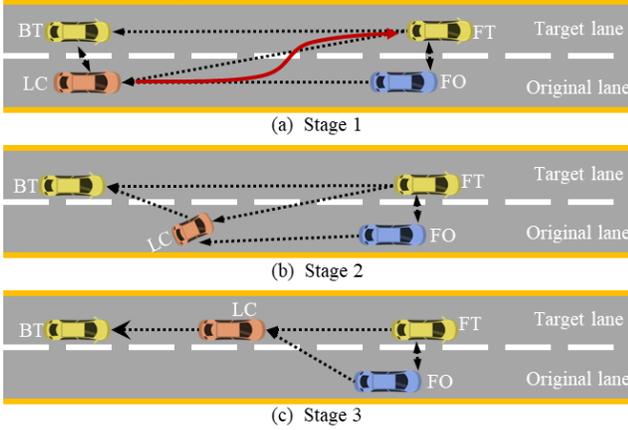

**Fig. 3.** Relationships Between SVs During Lane-Changing Process

This study focuses on the influence of SV behavior before and during a lane change on the interaction risk faced by the LCV at the time of completion. The vehicle behaviors and distances in Stage 1 and Stage 2 are selected as causal factors, as shown in Table 1 and Table 2, respectively. Stage 3 risk is defined as the maximum IS between the LCV and the SVs.

**Table 1** Causal factors selection for stage 1.

| Location | Influence factors | Abbreviation |
|---|---|---|
| Vehicles on the original lane | Speed and acceleration of LCV | lc_vx1, lc_vy1, lc_ax1, lc_ay1 |
| | Speed and acceleration of the front vehicle | fo_vx1, fo_vy1, fo_ax1, fo_ay1 |
| | Distance between the LCV and the front vehicle | d_lc_fo1 |
| Vehicles on the target lane | Speed and acceleration of the front vehicle | ft_vx1, ft_vy1, ft_ax1, ft_ay1 |
| | Speed and acceleration of the behind vehicle | bt_vx1, bt_vy1, bt_ax1, bt_ay1 |
| | Longitudinal distance between the LCV and the front vehicle | d_lc_ftx1 |
| | Lateral distance between the LCV and the front vehicle | d_lc_fty1 |
| | Longitudinal distance between the LCV and the behind vehicle | d_lc_btx1 |
| | Lateral distance between the LCV and the behind vehicle | d_lc_bty1 |
| | Distance between the front and behind vehicles | d_ft_bt1 |

**Table 2** Causal factors selection for stage 2.

| Location | Influence factors | Abbreviation |
|---|---|---|
| Vehicles on the original lane | Longitudinal and lateral speed and acceleration of the LCV | lc_vx2, lc_vy2, lc_ax2, lc_ay2 |
| | Speed of the front vehicle | fo_vx2, fo_vy2 |
| | Distance between the LCV and the front vehicle | d_lc_fo2 |
| Vehicles on the target lane | Longitudinal and lateral speed and acceleration of the front vehicle | ft_vx2, ft_vy2, ft_ax2, ft_ay2 |
| | Longitudinal and lateral speed and acceleration of the behind vehicle | bt_vx2, bt_vy2, bt_ax2, bt_ay2 |
| | Distance between the LCV and the front vehicle | d_lc_ftx2, d_lc_fty2 |
| | Distance between the LCV and the behind vehicle | d_lc_btx2, d_lc_bty2 |
| | Distance between the front and behind vehicles | d_ft_bt2 |

*B. Causal Inference of Interaction Risk for LCV*

The theory of event chain proposed by Heinrich indicates that accidents are triggered by a chain of causal events occurring in temporal order [32-34]. Based on the stage division of the lane-change process, this study estimates the causal effects between vehicle behaviors and driving risk through a causal pathway from initial conditions, through vehicle behaviors, to the final risk state, as illustrated in Fig. 3.

In Fig. 3, Stage 1 (time $t_1$) represents the initial conditions when the vehicle prepares to change lanes, including its longitudinal and lateral behaviors as well as spatiotemporal distances to SVs. Stage 2 (time $t_2$) corresponds to the lane-change process itself. Stage 3 (time $t_3$) represents the moment when the vehicle has completed the lane change. In this study, both the LCV and SVs are considered. The initial conditions at $t_1$ influence the behaviors of the lane-changing and SVs during $t_2$. In turn, the behaviors during $t_2$ affect the risk faced by the LCV at the completion of the maneuver.

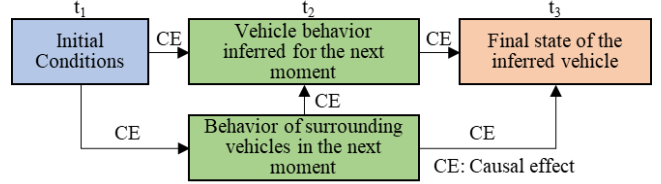

**Fig. 4** Framework for Inferring Causes of Traffic Risks

*a). Causal Graph*

Similar to previous studies [23, 35], the causal graph in this study is derived from expert prior knowledge. Fig. 5 illustrates the causal logic between vehicle behaviors and the final interaction risk. Distance refers to the longitudinal and lateral distance between the current vehicle and SVs. The current vehicle's speed, together with its distance to SVs, influences the driver's acceleration decision, which in turn affects the vehicle's speed and acceleration in Stage 2. As the calculation of interaction intensity considers vehicle speed, acceleration, and distance, the interaction intensity is also affected by these behaviors.

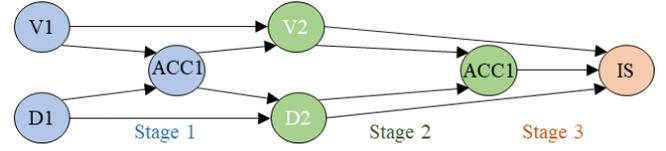

**Fig. 5.** Causal Logic Diagram of Single Vehicle and Final Interaction Risk.

Specifically, the LCV, FO, FT, and BT are considered together. In addition, in this study, the lateral and longitudinal behaviors of each vehicle are treated separately. The resulting causal graph between these factors and the final interaction risk of the LCV is shown in Fig. 6.

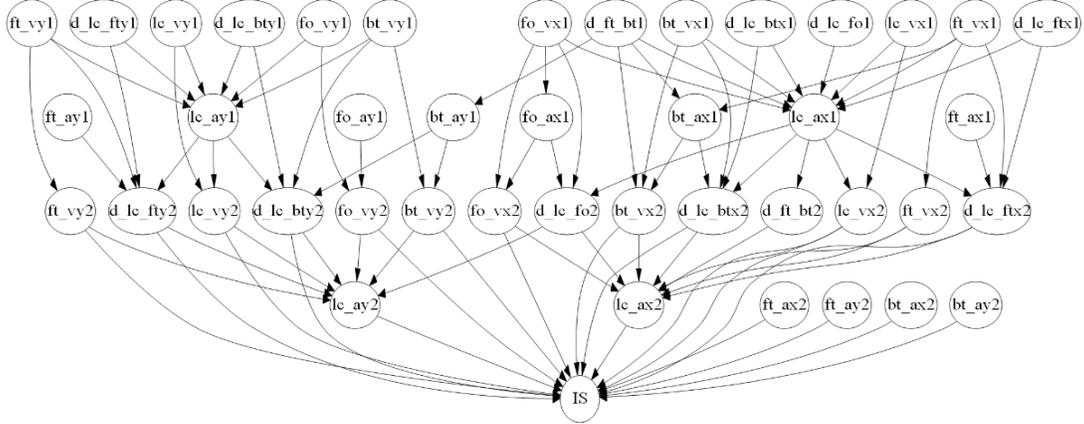

**Fig. 6.** Causal Diagram of LCV Interaction Risk.

*b). Causal Effect*

To verify the stability and interpretability of the causal effect estimation, this study first computes the average treatment effect (ATE) between intervention and outcome variables, followed by the CATE. CATE is the ATE of a subgroup in the population, and the conditional variable is the variable that divides the group. In this study, the ATE is used to check whether the CATE aligns with the global feature, while the CATE provides heterogeneous prior knowledge for the trajectory planning of LCVs.

(1) *Average treatment effect (ATE)*

The average causal effect is used to explain the mean impact of explanatory variables on the outcome variable, mainly by comparing the causal effect under different levels of the intervention variable. The treatment effect is an important tool for quantifying causal relationships and can be estimated based on treatment variables and outcome variables. In this study, the treatment variables refer to microscopic behaviors of vehicles, while the outcome variable is defined as the maximum IS between the LCV and SVs when the change of lanes is finished. ATE is calculated by:

$$E(Y|W, T = t_1) - E(Y|W, T = t_0) \quad (5)$$

where $W$ denotes the set of variables, including vehicle speed, acceleration, and other behaviors listed in Tables 1 and 2. $T$ represents the treatment variable, that is, the microscopic behavior of the vehicle selected from $W$. $Y$ is the outcome variable observed and measured by varying the treatment variable, defined as the maximum IS between LCV and SVs in Stage 3.

The key to calculating the ATE lies in estimating $E(Y|W, T)$. In causal inference, $E(Y|W, T)$ can be achieved through regression models, computed as follows:

$$E(Y|W, T) = \beta_0 + \beta_t T + \beta_w W \quad (6)$$

where $\beta_t$ average treatment effect. In the calculation of causal effects, the generalized propensity score (GPS) [36] can be employed for estimation. Then, $E(Y|W, T)$ can be expressed as:

$$E(Y|W, T) = \beta_0 + \beta_t F(T) + \beta_w R(T, W) \quad (7)$$

where $R(T, W)$ denote the generalized propensity score, and $F(T)$ be a constructed function of the treatment variable. Assuming $T_i | W_i \sim N(\alpha_0 + \alpha_1 W_i, \sigma^2)$, $(\alpha_0, \alpha_1, \sigma^2)$ can be estimated via maximum likelihood, $R(T, W)$ can be expressed as:

$$\hat{R}_i(T_i, W_i) = \frac{1}{\sqrt{2\pi\hat{\sigma}^2}} \exp\left(-\frac{1}{2\hat{\sigma}^2}(T_i - \hat{\alpha}_0 - \hat{\alpha}_1 W_i)\right) \quad (8)$$

$Y_i$ can be predicted by treating $T_i$ and $R_i$ as independent variables. $(\hat{\beta}_0, \hat{\beta}_t, \hat{\beta}_w)$ can be estimated using a regression model, which can be expressed as:

$$E(Y|W, T) = \beta_0 + \beta_1 T + \beta_2 T^2 + \beta_3 R + \beta_4 R^2 + \beta_5 TR \quad (9)$$

By this regression model, $(\hat{\beta}_0, \hat{\beta}_1, \hat{\beta}_2, \hat{\beta}_3, \hat{\beta}_4, \hat{\beta}_5)$ can be estimated. Therefore, the treatment effect can be estimated as:

$$\hat{Y}(T = T_i, W = W_i) = \hat{\beta}_0 + \hat{\beta}_1 T + \hat{\beta}_2 T^2 + \hat{\beta}_3 \hat{R}_i(T_i, W_i) + \hat{\beta}_4 \hat{R}_i(T_i, W_i)^2 + \hat{\beta}_5 T \hat{R}_i(T_i, W_i) \quad (10)$$

**(2)** *Conditional average treatment effect (CATE)*

Under naturalistic driving conditions, drivers exhibit heterogeneity [37-39]. To further quantify the heterogeneity of different drivers during lane-changing maneuvers and provide personalized prior knowledge for vehicle trajectory planning, the double machine learning (DML) was employed in this study[23].

The CATE represents the causal effect within a subpopulation defined by specific conditions. The input variables for CATE estimation include the treatment variable (e.g., SV speed), the outcome variable (e.g., risk during phase 3 of lane-changing), and the conditioning variables (e.g., other influencing factors).

In this study, tree-based models are employed to estimate the causal effect of treatment variables (i.e., treatment effects). The conditional regression function can be expressed as:

$$\mu_t(X) = E[Y(t)|X] \quad (11)$$

where $\hat{\mu}_t(X) = \mu_t(X, \hat{W}_t) = E[Y(t)|X, W_t]$, $X$ is treatment variable, $T$ denotes the treatment variable, and $Y(t)$ represents the outcome variable under the condition $T=t$.

To compute the CATE, this study employs gradient boosting regression to fit $\mu(X,W) = E[Y|X,W]$ and $f(X,W) = E[T|X,W]$, thereby obtaining local estimates of $\hat{\mu}_i(X,W)$, and $\hat{f}(X,W)$.

Furthermore, to avoid issues caused by ill-conditioned local matrix equations, which may lead to inaccurate or unreliable parameter estimation, causal tree models [39] are adopted in the estimation process, i.e.,

$$E\begin{bmatrix}(Y - E[Y|X,W] - \theta(x)(T - E[T|X,W])) \\ \times (T - E[T|X,W])|X = x\end{bmatrix} = 0 \quad (12)$$

In the estimation process, double machine learning (DML) [40] is employed to compute the CATE. The core idea of DML is to use two independent machine learning models to estimate causal effects, where one model is employed to correct the bias of the other, leading to more accurate estimation. The objective function of DML is given as:

$$\hat{\theta}(x) = \arg\min_\theta \sum_{i=1}^n K_x(X_i) \cdot (Y_i - \mu_i(X_i, W_i) \\ - \theta(T_i - \hat{f}(X_i - W_i)))^2 \quad (13)$$

where $K_x(X_i)$ denotes the similarity matrix.

To validate the estimation results, this study adopts a placebo test by refitting the model [41]. Specifically, the treatment variable is replaced with random data independent of other variables, and the treatment effect is recalculated. The validity of the estimation is then assessed by comparing the new treatment effect with the original one. If the placebo replacement generates a treatment effect with a different distribution, the original treatment effect is considered valid.

C. Trajectory planning for lane changing vehicle

In this section, the causal effects among vehicle behaviors are incorporated into the lane-change trajectory planning method to improve the performance of the trajectory planner. Firstly, the vehicle dynamics model is introduced. Then, the MPC-based trajectory planning process for LCVs is described. Finally, causal prior knowledge is integrated into the trajectory planning method.

*a). Vehicle Module*

Fig. 7 shows the vehicle dynamics model. The control inputs are the front-wheel steering angle δ and the acceleration $a$. The state $s_e$ and $u_e$ of the LCV are defined as:

$$s_e = [x_e, y_e, \upsilon_e, \varphi_e] \quad (14)$$
$$u_e = [a_e, \delta_e] \quad (15)$$

where $(x_e, y_e)$ denote the coordinates of the rear wheel position; $\varphi$ represents the yaw angle; $v_e$ and $a_e$ denote the vehicle speed and acceleration, respectively; and $\delta_e$ denotes the steering angle.

The vehicle center is located at the midpoint of the rear axle. Therefore, the kinematic model of the vehicle can be expressed as [42]:

$$\dot{x}_e = v_e \cos(\varphi_e) \quad (16)$$
$$\dot{y}_e = v_e \sin(\varphi_e) \quad (17)$$
$$\dot{v}_e = a_e \quad (18)$$
$$\dot{\varphi}_e = \frac{v_e \tan(\delta_e)}{L_e} \quad (19)$$

where $L_e$ denotes the wheelbase of the vehicle. The predicted vehicle states are calculated using Eq. (20), and further expanded with a Taylor series to obtain Eq. (21) [21].

$$s_e(t+1) = s_e(t) + \dot{s}_e \Delta t \quad (20)$$
$$s_e(t+1) = As_e(t) + Bu_e(t) + C \quad (21)$$

where:

$$A = \begin{bmatrix} 1 & 0 & \cos(\varphi_e)\Delta t & -v\sin(\varphi_e)\Delta t \\ 0 & 1 & \sin(\varphi_e)\Delta t & v\cos(\varphi_e)\Delta t \\ 0 & 0 & 1 & 0 \\ 0 & 0 & \frac{v_e \tan(\delta_e)}{L_e}\Delta t & 1 \end{bmatrix} \quad (23)$$

$$B = \begin{bmatrix} 0 & 0 \\ 0 & 0 \\ \Delta t & 0 \\ 0 & \frac{v_e}{L_e \cos^2(\delta_e)}\Delta t \end{bmatrix} \quad (24)$$

$$C = \begin{bmatrix} v_e \sin(\varphi_e)\varphi_e \Delta t \\ -v_e \cos(\varphi_e)\varphi_e \Delta t \\ 0 \\ \frac{v_e \delta_e}{L_e \cos^2(\varphi_e)}\Delta t \end{bmatrix} \quad (25)$$

*b). MPC Implementation*

MPC-based vehicle trajectory planning is a model-based control approach. The vehicle dynamic model is used to predict its behavior over a future time horizon, and these predictions are then used to generate an optimal trajectory. During this process, MPC considers multiple constraints, including vehicle dynamic constraints, environmental constraints (e.g., collision avoidance), and path constraints, to ensure that the generated trajectory is both safe and feasible. The MPC framework is illustrated in Fig. 7.

Firstly, the vehicle's initial position, speed, steering angle, and target position are obtained from real trajectory data. Next, a reference trajectory for the LCV is generated based on the potential field distribution of SVs, as shown in Fig. 8, from which the corresponding speed and steering angle are derived. Then, MPC is executed to search for the optimal trajectory, steering angle, and speed. In this process, the vehicle kinematics serve as constraints, and the cost function accounts for control effort, state deviation, control rate change, and terminal state error. Finally, the control output for the first time step is selected, representing the actual trajectory to be executed. This procedure is iterated to ensure that the LCV reaches its destination smoothly and safely, thereby completing the lane-change maneuver.

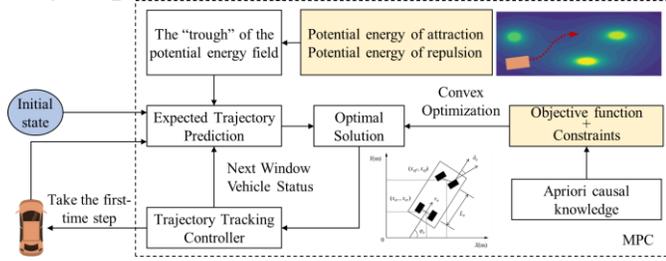

**Fig. 7.** The trajectory planning framework based on PF and MPC.

In the potential field-based trajectory planning stage, the destination of LCV is first set. The repulsive potential from SVs and road edges, as well as the attractive potential from the destination, are then calculated for the LCV. The repulsive potential is computed as follows:

$$E_{repulsive} = \sum_{j=1}^{S_{num}} EPE_j^e \quad (26)$$

where $EPE_j^e$ represents the potential generated by the j-th SV, calculated as described in [25]. $S_{num}$ denotes all vehicles surrounding the LCV. In this study, it includes FO, FT, and BT, as shown in Fig. 3. The attractive potential from the destination is calculated as follows:

$$E_{attractive} = K_p \sqrt{(x - x_{des})^2 + (y - y_{des})^2} \quad (27)$$

where $(x_{des}, y_{des})$ denotes the coordinates of the local lane-change destination. $K_p$ is the scaling factor for the attractive potential, set to 20 in this study. The overall potential field can then be expressed as:

$$E_{sum} = E_{repulsive} + E_{attractive} \quad (28)$$

The characteristics of the potential field indicate that the total potential experienced by the LCV should be minimized to ensure a safe lane change [43]. Therefore, the vehicle's local reference trajectory is defined as the point corresponding to the minimum of $E_{sum}$, i.e., the "valley" of the potential field.

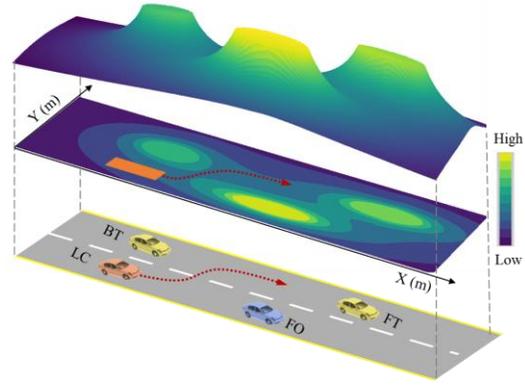

**Fig. 8.** Schematic Diagram of Potential Energy Field Trajectory Search.

After initialization, a reference state is provided to the MPC controller to perform local path planning and tracking. First, the vehicle motion space is initialized and discretized to obtain all possible movements. Second, the discrete set of points corresponding to the minimum potential in the valley is computed. Then, a spline curve [44] is applied to connect these discrete points into a continuous curve, yielding the reference path ($x_{ref}$, $y_{ref}$, $\varphi_{ref}$). Finally, a set of reference states [$x_{ref}$, $y_{ref}$, $v_{ref}$, $\varphi_{ref}$] is generated. Once the reference states are obtained, the objective is to minimize the error between the vehicle states and the references:

$$J_1 = x_{err}(t_i) = \sum_{i=1}^{N_p} (x(t_i) - x_{ref}(t_i))^2 \quad (29)$$

$$J_2 = y_{err}(t_i) = \sum_{i=1}^{N_p} (y(t_i) - y_{ref}(t_i))^2 \quad (30)$$

$$J_3 = v_{err}(t_i) = \sum_{i=1}^{N_p} (v(t_i) - v_{ref}(t_i))^2 \quad (31)$$

$$J_4 = \varphi_{err}(t_i) = \sum_{i=1}^{N_p} (\varphi(t_i) - \varphi_{ref}(t_i))^2 \quad (32)$$

where, $N_p$ is the prediction horizon. $x_{err}(t_i)$, $y_{err}(t_i)$, $v_{err}(t_i)$ and $\varphi_{err}(t_i)$ denote the relative errors of the lateral position, longitudinal position, speed, and heading angle of the LCV at prediction time $t_i$. $x(t_i)$, $y(t_i)$, $v(t_i)$ and $\varphi(t_i)$ represent the predicted lateral position, longitudinal position, speed, and heading angle of the LCV at time $t_i$.

The LCV travels along the "valley" of the safety field through $J_1$ and $J_2$. To reduce sharp steering, sudden acceleration, and abrupt deceleration, thereby improving the stability and comfort of the lane change, variations in steering angle and acceleration are considered.

$$J_5 = \sum_{j=0}^{N_c} (a(t_{j+1}) - a(t_j))^2 \quad (33)$$

$$J_6 = \sum_{j=0}^{N_c} (\delta(t_{j+1}) - \delta(t_j))^2 \quad (34)$$

where, Nc denotes the maximum control horizon. $a(t_j)$ and $\delta(t_j)$ represent the acceleration and steering angle at the control time step $t_j$, respectively. In addition, constraints are applied to the optimization problem to ensure that the vehicle

does not exceed its kinematic limits. The constraints are expressed as follows:

$$\min_{u,\varepsilon} \sum_{k=1}^{6} \tau_k J_k + \|\varepsilon_t\|^2 \quad (35)$$

$$\text{sub. to } s(t+1) = As(t) + Bu(t) + C \quad (36)$$

$$0 \leq \delta(t_j) \leq \delta_{max} \quad (37)$$

$$-a_{dec} \leq a(t_j) \leq a_{acc} \quad (38)$$

$$-\varphi_{max} \leq \varphi(t_i) \leq \varphi_{max} \quad (39)$$

$$0 \leq v(t_i) \leq v_{max} \quad (40)$$

where, $\tau_k$ is the weighting parameter, where k ranges from 1 to 6. $\varepsilon_t$ denotes the vectorized slack variable for the past $t$ steps at the current time. $\delta_{max}$ is the maximum steering angle; $a_{acc}$ and $a_{dec}$ are the maximum acceleration and deceleration, respectively. $\varphi_{max}$ is the maximum yaw angle, and $v_{max}$ is the maximum speed. At each sampling step, the optimal control inputs are obtained by solving the optimization problem in Eq. (35). The vehicle states are then computed by substituting the control inputs into Eq. (21). Repeating this procedure yields the dynamically planned vehicle trajectory.

*c). Embedding Priori Causal Knowledge into MPC*

In this study, the causal effects between lane-change risk and the behaviors of SVs are incorporated into the objective function of MPC. During the lane-change process, only the LCV exhibits significant lateral velocity. Therefore, the cost considers the longitudinal speed difference between the LCV and SVs, with larger differences resulting in higher costs.

In addition, the spatial distance between the LCV and SVs should also be considered. Specifically, the LCV should maintain as large a distance as possible from nearby vehicles during the maneuver. However, keeping a larger distance from one vehicle may reduce the distance to another. For example, if the LCV maintains a large longitudinal distance from the leading vehicle in the target lane, its longitudinal distance to the following vehicle in the same lane will become smaller.

Accordingly, two additional terms are included in the trajectory planning cost function: the longitudinal distance and the speed difference between the LCV and SVs. The cost weights for different SVs are set in accordance with the estimated causal effects. The modified MPC objective function is expressed as:

$$\min_{u,\varepsilon} \sum_{k=1}^{6} \tau_k J_k + \sum_{i=1}^{3} \alpha_i (x(t_i) - xo(t_i))^2 \\ + \sum_{i=1}^{3} \beta_i (v(t_i) - vo(t_i))^2 + \|\varepsilon_t\|^2 \quad (41)$$

where $xo(t_i)$ and $vo(t_i)$ denote the longitudinal position and longitudinal speed of the SVs at time $t_i$. The SVs include three vehicles, as shown in Fig. 3. $\alpha_i$ and $\beta_i$ are the weights for longitudinal distance difference and speed difference with respect to different SVs.

In particular, the values of $\alpha_i$ and $\beta_i$ are determined by the estimated causal effects. Specifically, the initial conditions before the lane change are used as inputs, and the corresponding CATE values are used as outputs to train a machine learning model. During new lane-change trajectory planning, the new initial conditions are used as inputs, and the predicted causal effects among behaviors are taken as weight values in the modified objective function. The lane-change is then executed following the workflow illustrated in Fig. 7.

## III. CASE STUDY

### A. Empirical Data

The datasets of HighD [45] and CitySim [46] are employed to extracts vehicle trajectory used in this study. Each lane-change sequence includes the complete process before, during, and after the maneuver, with a total of 880 lane-change cases selected.

Both HighD and CitySim datasets are derived from UAV-recorded traffic videos, from which vehicle trajectories are obtained using image recognition techniques. The trajectory data include longitudinal and lateral positions, speed, and acceleration. Vehicle position data are measured in meters, while speed and acceleration are measured in m/s and m/s², respectively. The data frequency is uniformly 5 Hz. These datasets have already been widely applied in studies on microscopic vehicle behavior [47-50].

### B. Causal Inference Results

*a). Results of ATE*

Fig. 9 illustrates the causal effects of lateral and longitudinal speeds of different vehicles on the interaction risk at the completion of a lane change, along with the corresponding placebo effects. The red solid dots represent the ATE, while the gray boxplots show the distribution of causal effects from 50 randomly generated placebo datasets. It can be observed that the average causal effects of the variables are significant, whereas the placebo effects are close to zero, indicating the stability of the estimated causal effects.

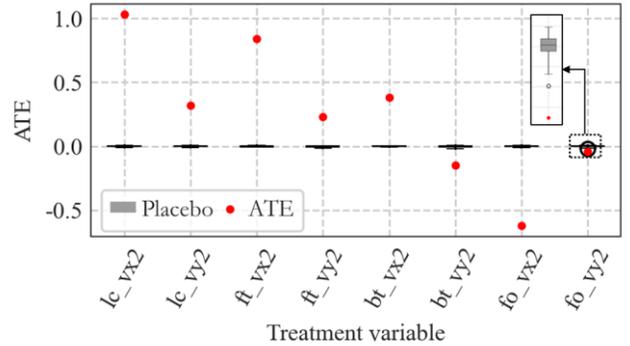

**Fig. 9.** Causal Effect Distribution of Lateral and Longitudinal Velocity Among Different Vehicles.

In Fig. 12, the average causal effects of 1 lc_vx2, lc_vy2, ft_vx2, ft_vy2, bt_vx2, bt_vy2, fo_vx2 and fo_vy2 are 1.03, 0.32, 0.84, 0.23, 0.38, –0.15, –0.62, and –0.04, respectively.

The lateral and longitudinal speeds of the LCV have positive causal effects on the final interaction risk, indicating that higher speeds during the lane change lead to greater interaction risk. Similarly, a higher lateral speed of the leading vehicle in the target lane may increase the final interaction risk. This is because faster leading vehicles exert influence over a larger area, causing higher interaction intensity after the LCV merges. The lateral and longitudinal speeds of the following vehicle in the target lane also positively affect the final interaction risk for similar reasons.

In addition, the lateral and longitudinal speeds of the leading vehicle in the original lane have negative causal effects on the final interaction risk. This indicates that faster vehicles in the original lane reduce the final risk of the LCV in the target lane. This is reasonable because faster vehicles in the original lane provide more available space for the LCV, allowing the driver to select a safer merging state.

*b). Results of CATE*

During natural driving, drivers determine their next actions based on the current vehicle speed and the distance to SVs that may potentially lead to conflicts. Therefore, for the eight vehicle behavior variables shown in Fig. 12, the conditional individual causal effects are calculated. The conditional variables selected for each factor are listed in Table 3.

**Table 3** Conditional variable selection for different vehicle behavior factors.

| Treatment variables | lc_vx2, lc_vy2, ft_vx2, ft_vy2, bt_vx2, bt_vy2, fo_vx2, fo_vy2 |
|---|---|
| Conditional variable | d_ft_bt2, d_lc_bty2, d_lc_ftx2, d_lc_fty2, d_lc_btx2, d_lc_bty2, d_lc_fo2, d_lc_fo2 |

Fig. 10 shows the probability distributions of the individual causal effects for the eight variables, with mean values of 0.56, 0.36, 0.93, 0.12, 1.25, –0.47, –0.2, and 0.0, respectively. Compared to the Average Treatment Effect (ATE), the CATE captures individual-level causal effects, which helps reveal driver heterogeneity. For example, the distribution of lc_vx2 is mainly on the positive side, consistent with the sign of the ATE, indicating that the causal direction estimated by both methods aligns. However, a portion of the causal effects lies on the negative side, suggesting that some drivers, even at higher speeds, can merge into the target lane with lower interaction risk. This may reflect higher driving skills, enabling them to select safer merging states.

It is noteworthy that in the CATE results, the causal effect of fo_vy2 is zero, and no heterogeneity is observed among individuals. A possible reason is that, as the LCV moves away from the original lane, the lateral speed of the leading vehicle in the original lane does not causally affect the final interaction risk of the LCV. Such an effect cannot be captured by the average causal effect. However, DML, by leveraging both the outcome model $Y$ and the treatment model $T$, can better assign units to the treatment and control groups, providing an unbiased estimate of the causal effect.

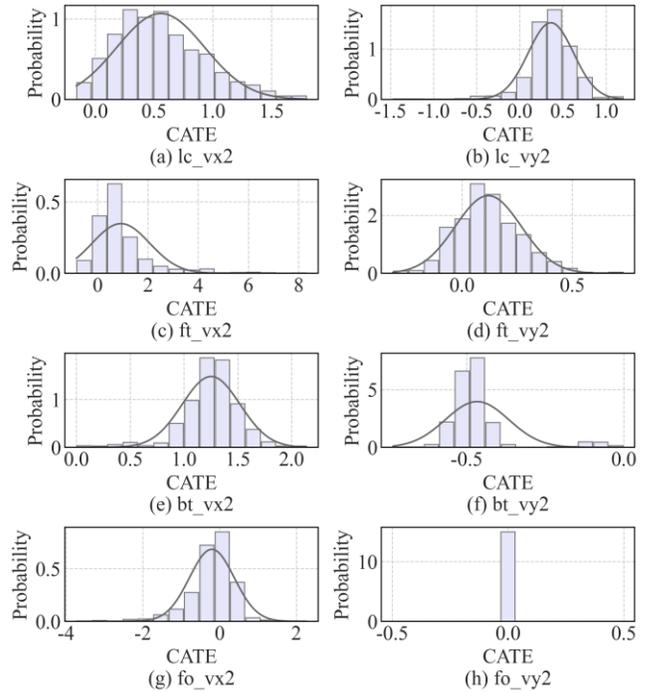

**Fig. 10.** Probability distribution of CATE for eight variables during lane changes.

To compare the reproducibility of individual causal effects across different machine learning models, support vector machine (SVM) [51], random forest (RF) [39], and XGBoost [52] models were trained. Seventy-five percent of the data were used as the training set, and 25% as the test set. The Root Mean Square Error (RMSE) was calculated for all three models. The results indicate that XGBoost achieved the best performance.

*C. Lane Changing Vehicle Trajectory Planning*

Before planning the trajectory of the LCV, the trajectory information of the LCV and the three SVs at the initial time step is selected for initialization. The trajectories of the SVs are then separated and used to update the obstacle trajectories during the LCV's trajectory planning. The simulation setup is summarized in Table 4.

**Table 4** MPC Initial Parameter Settings.

| Parameters | Setting |
|---|---|
| Desired speed | $v_{last}$ |
| Maximum reverse speed | 0 |
| Maximum steering angle | $\pi/4$ |
| Maximum steering rate | $\pi/6$ |
| Maximum acceleration | 6 m/s$^2$ |
| Simulation step length | 5 |
| Time step | 0.2 s |

Fig. 11 illustrates an example of vehicle trajectory planning, with the lane-changing duration of approximately 5 s. In Fig. 11(a), the black solid line represents the planned trajectory of the LCV, while the dark red solid lines denote the trajectories of SVs. It can be observed that the planned trajectory enables a relatively smooth insertion from the original lane into the target lane. Fig. 11(b) shows the heading angle of the LCV. At the initial stage of the maneuver, the vehicle heading exhibits a larger deviation, which is then gradually adjusted during the insertion process. Overall, the heading angle remains relatively small.

Fig. 11(c) and 12(d) present the longitudinal and lateral speed of the LCV, respectively. The longitudinal speed remains relatively stable throughout the maneuver. The lateral speed first increases and then decreases, which resembles the typical lane-changing behavior of human drivers. Specifically, the lateral velocity increases during the initiation phase of the maneuver, and gradually decreases as the vehicle approaches the target lane, eventually approaching zero once the maneuver is completed.

Fig. 11(e) depicts the acceleration of the LCV. Under the MPC-based planning, rapid accelerations or decelerations (absolute values greater than 3 m/s²) are rarely observed. Nevertheless, the planned acceleration still exhibits noticeable fluctuations overall.

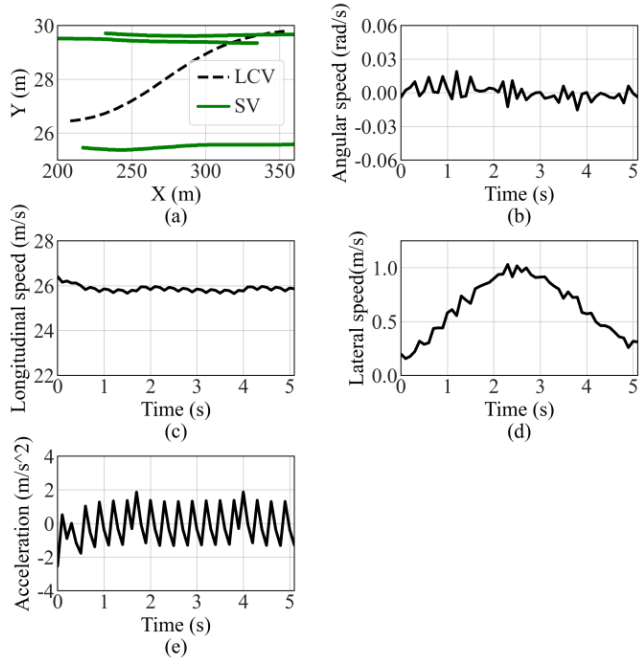

**Fig. 11.** Examples of planning results: (a) LCV trajectory, (b) Heading angle, (c) Longitudinal speed, (d) Lateral speed, (e) Acceleration.

Fig. 12 presents the boxplots of trajectory and lateral speed deviations between the planned and actual trajectories for the original MPC and the modified MPC. As shown in Fig. 12(a), the trajectory deviations of the original MPC are significantly larger, with a maximum deviation of approximately 1.2 m. In contrast, the deviations obtained from the proposed MPC are substantially smaller, all below 0.2 m.

Fig. 12(b) illustrates the deviations in lateral LCV. The original MPC exhibits much larger discrepancies, with a maximum deviation of about 7 m/s and an average deviation of around 0.3 m/s, which is clearly unreasonable. In comparison, the proposed MPC achieves a maximum deviation of only about 1 m/s, with an average deviation of approximately 0.12 m/s. The smaller lateral speed deviations indicate that the proposed MPC enables smoother and more reasonable insertion into the target lane, thereby improving driving comfort.

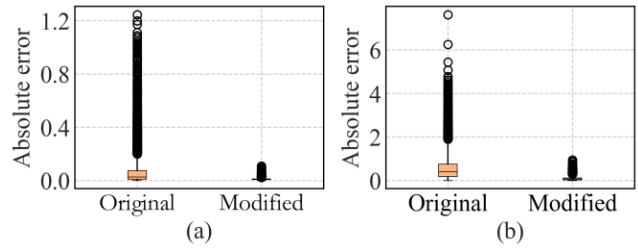

**Fig. 12.** Box plot of differences for (a) trajectory coordinate, and (b) lateral speed.

Fig. 13 presents the comparison between the original MPC and the modified MPC in terms of vehicle trajectory and lateral speed. As shown in Fig. 13(a), the trajectory generated by the modified MPC is smoother, while the trajectory obtained from the original MPC exhibits significant fluctuations, which can negatively affect driving comfort. Similarly, the lateral velocity planned by the original MPC shows greater magnitude and variability compared with the modified MPC.

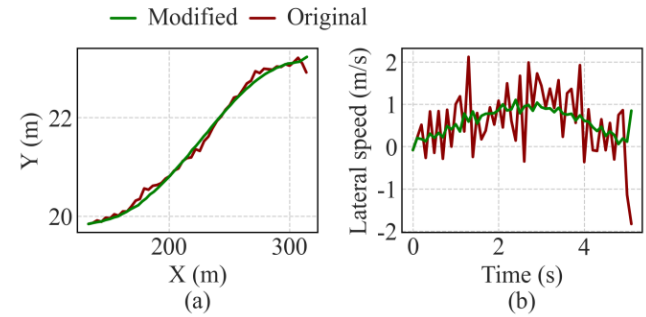

**Fig. 13.** Example of comparison between original MPC and MPC results in this paper: (a) trajectory points, (b) lateral speed.

Fig. 14 compares the longitudinal speed and acceleration between the original MPC and the proposed MPC. As shown in Fig. 14(a), the difference in longitudinal speed is negligible, and both approaches maintain stable velocity profiles. However, Fig. 14(b) reveals that the acceleration obtained from the original MPC exhibits significantly larger fluctuations, leading to frequent rapid acceleration and deceleration. Such instability not only reduces driving comfort but may also pose safety risks in complex traffic environments. In contrast, the modified MPC produces smoother acceleration profiles, thereby enhancing both comfort and safety during lane changes.

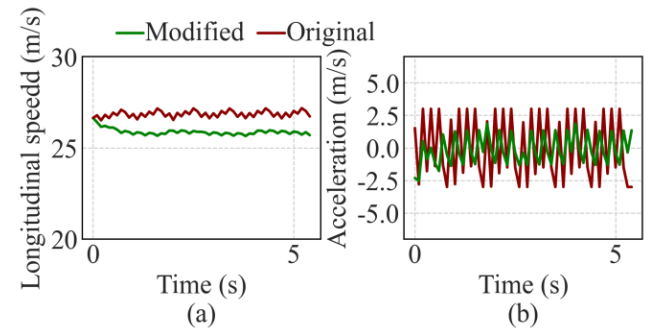

**Fig. 14.** Comparison of original MPC and modified MPC results, (a) Longitudinal speed, (b) Longitudinal acceleration.

Fig. 15 illustrates the difference in yaw angle between the original MPC and the modified MPC. The overall trend shows that the yaw angle is relatively large at the beginning of the lane-change maneuver, then remains mostly stable, with several adjustments made before completing the insertion into the target lane. As we can see that the yaw angle planned by the original MPC fluctuates more drastically, exhibiting excessive corrections. Such instability can reduce driving comfort and pose safety concerns during the maneuver. In contrast, the modified MPC produces smoother yaw angle variations, better aligning with human driving behavior.

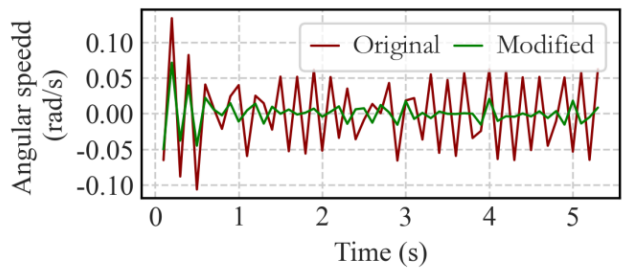

Fig. 15. Comparison of vehicle yaw angle between original MPC and improved MPC results.

## IV. DISCUSSION

This study quantified the causal relationships between vehicle-level microscopic behaviors, thereby revealing their mutual influences. To demonstrate that causal effect estimation can uncover more insightful results, we compared the differences between causal effects and correlation coefficients. Fig. 16 presents the correlation coefficients between the eight treatment variables and the interaction risk at the completion of lane changes. In most cases, the signs of the correlation coefficients are consistent with those of the causal effects. This consistency can be explained by the fact that the treatment variables were selected based on expert knowledge as potential causal factors, and all eight variables are directly connected to the final risk of lane changes in the causal graph.

Interestingly, the longitudinal velocity of the leading vehicle in the original lane (fo_vx2) shows a correlation coefficient with a sign opposite to that of the causal effect. When computing the correlation, only the direct relationship between the leading vehicle in the target lane and the final interaction risk was considered. However, in the true causal relationship, the longitudinal velocity of the leading vehicle in the target lane first affects the longitudinal acceleration of the LCV, which in turn influences the final risk. A higher speed of the leading vehicle provides the driver with more space to adjust the merging timing and position, thereby reducing the interaction risk. Consequently, the relationships between vehicle behaviors and risk derived from the causal graph align better with human driving cognition.

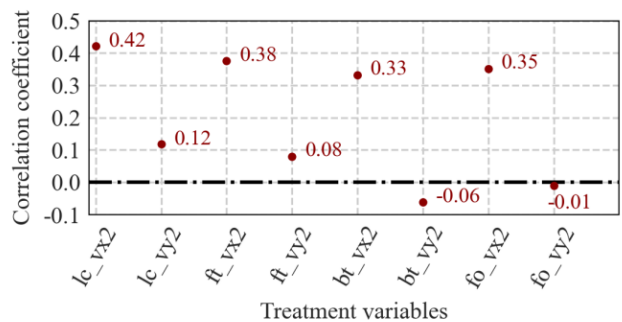

**Fig. 16.** Scatter plot of correlation coefficients between intervention variables and final interaction risk of LCVs.

## V. CONCLUSION

Lane-change trajectory planning has long been a key focus in autonomous driving research. In this study, the temporal process of lane-changing maneuvers was considered, and a method for computing causal effects among vehicle microscopic behaviors was developed. Furthermore, a lane-changing trajectory planning method based on MPC incorporating causal priors was proposed. The results demonstrate that the causal effects computed using causal inference techniques are highly interpretable. Compared with the baseline MPC, the maximum trajectory deviation decreased from 1.2 m to 0.2 m, the lateral velocity fluctuation reduced from 0.3 m/s to 0.12 m/s, the yaw angle variability was reduced by 50%, and longitudinal acceleration fluctuations were also smaller.

The proposed lane-changing trajectory planning method with causal priors can be applied to both autonomous driving capability enhancement and autonomous vehicle simulation testing. On the one hand, the integration of causal prior knowledge into trajectory planning improves the human-likeness of lane-changing maneuvers [53, 54]. Moreover, the proposed framework can be readily extended to other scenarios, such as trajectory planning in weaving areas. On the other hand, the method provides potential support for autonomous vehicle simulation testing [55-57]. For instance, during test scenario generation, the proposed approach can be used to plan the formal trajectories of background vehicles, thereby enabling natural interactions and adversarial behaviors between the test vehicle and surrounding vehicles.

This study explores the integration of causal inference with MPC, providing a new approach to improve both safety and comfort in autonomous driving. Future work should further investigate several aspects. First, methods for automatically generating causal graphs as prior knowledge need to be developed. Second, incorporating prior knowledge into more advanced models, including reinforcement learning, remains a promising direction. Finally, extending the application of this approach to more complex driving tasks, such as behavioral decision-making, warrants further exploration.

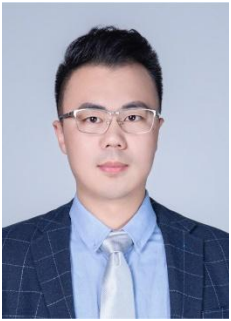

**Cailin Lei** received the Ph.D. degree with the College of Transportation, Tongji University, China, in 2024. He is currently Assistant Professor with the School of Smart City, Chongqing Jiaotong University. His research interests include cooperative vehicle infrastructure systems, data mining, driving behavior analysis and behavioral decision-making of intelligent vehicles.

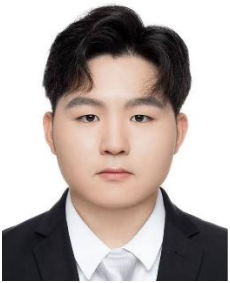

**Haiyang Wu** received the Master's degree in Electronic Information from Xijing University, Xi'an, Shaanxi, China, in 2025. He is currently pursuing the Ph.D. degree with the School of Transportation Engineering, Chongqing Jiaotong University. His research interests include autonomous driving, object detection, and intelligent transportation systems.

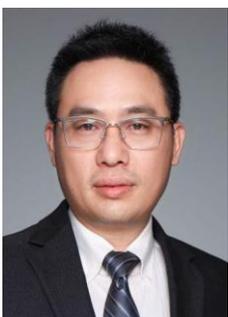

**Yuxiong Ji** received the B.S. and M.S. degrees in transport engineering from Tongji University in 2001 and 2004, respectively, and the Ph.D. degree in civil engineering from The Ohio State University, Columbus, OH, USA, in 2011. He is currently a Professor with the College of Transportation Engineering, Tongji University. His research interests include trustworthy testing of autonomous driving, cooperative vehicle infrastructure systems, data mining, and traffic operation and control.

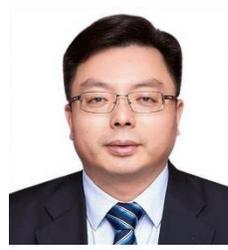

**Xiaoyu Cai** received his B.E. and M.E. degrees in Highway & Urban Road and Railway Engineering from Chongqing Jiaotong College in 2001 and 2004 respectively, and his Ph.D. degree in Road and Railway Engineering from Tongji University in 2007. He is currently a Professor at the School of Smart City, Chongqing Jiaotong University. His research focuses on traffic fusion perception and digital twins, intelligent monitoring and early warning for digital transportation systems, and development of intelligent transportation applications.

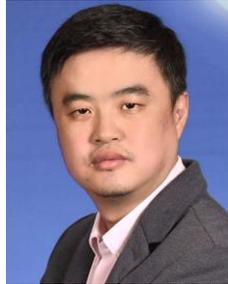

**Yuchuan Du** (Member, IEEE) received the B.S. and M.S. degrees in road engineering and the Ph.D. degree in traffic engineering from Tongji University, Shanghai, China, in 1998, 2001, and 2004, respectively. From 2003 to 2006, he was an Assistant Professor with the College of Transportation Engineering, Tongji University, where he was an Associate Professor, from 2006 to 2010. He is currently a Professor with the College of Transportation Engineering, Tongji University. His current research interests include innovative technologies for smart transportation infrastructure and intelligent transportation systems.